\documentclass[preprint,12pt,3p]{elsarticle}

\usepackage{mathtools}
\usepackage{amssymb}
\usepackage{amsfonts}
\usepackage{lipsum}
\usepackage{graphicx}
\usepackage{epstopdf}
\usepackage[export]{adjustbox}
\usepackage{amsthm}
\usepackage[author={Sina}]{pdfcomment}
\usepackage{soul, color}
\usepackage{float}
\usepackage{placeins}
\usepackage{siunitx}
\usepackage{multirow}
\usepackage{array}
\usepackage[capitalise]{cleveref}
\usepackage{amsfonts}
\usepackage{amsmath}
\DeclareMathOperator*{\argmin}{argmin} 

\makeatletter
\def\blfootnote{\xdef\@thefnmark{}\@footnotetext}
\makeatother

\begin{document}

\begin{frontmatter}

\title{Physics-Informed Neural Network for Modelling the Thermochemical Curing Process of Composite-Tool Systems During Manufacture}

\author{Sina Amini Niaki\textsuperscript{\textit{a,b,*}}} 
\author{Ehsan Haghighat\textsuperscript{\textit{c}}}
\author{Trevor Campbell\textsuperscript{\textit{a,d}}}
\author{Anoush Poursartip\textsuperscript{\textit{b,e}}}
\author{Reza Vaziri\textsuperscript{\textit{b,e}}}
\address[label11]{Data Science Institute (DSI), The University of British Columbia}
\address[label2]{Composites Research Network (CRN), The University of British Columbia}
\address[label3]{Department of Civil Engineering, The University of British Columbia}
\address[label4]{Department of Statistics, The University of British Columbia}
\address[label5]{Department of Materials Engineering, The University of British Columbia}
\cortext[cor1]{Corresponding author (\emph{Email}: saniaki@alumni.ubc.ca)}

\begin{abstract}

We present a Physics-Informed Neural Network (PINN) to simulate the thermochemical
evolution of a composite material on a tool undergoing cure in an autoclave. 
In particular, we solve the governing coupled system of differential equations---including
conductive heat transfer and resin cure kinetics---by optimizing the parameters of a deep neural
network (DNN) using a physics-based loss function. 
To account for the vastly different behaviour of thermal conduction and resin cure,
we design a PINN consisting of two disconnected subnetworks, and develop a sequential
training algorithm that mitigates instability present in traditional training methods.
Further, we incorporate explicit discontinuities into the DNN at the composite-tool interface and enforce known physical behaviour directly in the loss function to improve the solution 
near the interface.
We train the PINN with a technique that automatically adapts the weights on the loss terms
corresponding to PDE, boundary, interface, and initial conditions. 
Finally, we demonstrate that one can include problem parameters as an input to the model---resulting in a surrogate
that provides real-time simulation for a range of problem settings---and that one can use transfer learning
to significantly reduce the training time for problem settings similar to that of an initial trained model.
The performance of the proposed PINN is
demonstrated in multiple scenarios with different material thicknesses and thermal
boundary conditions.
\end{abstract}


\end{frontmatter}

\section{Introduction}\label{sec:intro}
The manufacture of polymeric composite materials involves forming prepreg, e.g. carbon fibres in a resin system, on a tool followed by subjecting the composite-tool system to a controlled cycle of temperature inside an autoclave \cite{campbell2003manufacturing}. The elevated temperature within the autoclave cures the polymeric resin resulting in a final composite product. The physics of this thermochemical process is well-established and expressed by a set of coupled nonlinear partial differential equations (PDEs) describing heat conduction and resin cure kinetics \cite{johnston1996process, boyard2016heat}. However, the solution of these PDEs is not available in closed form; computational approximation is required. A popular approach is the  finite element method (FEM) \cite{zienkiewicz2005finite}, which (1) approximates the solution to the PDE at each point in time using a collection of basis functions within discretized sub-domains of the geometry, and (2) iteratively simulates the evolution of the basis function coefficients forward in time. This method has been used to solve the set of governing differential equations for various stages of composite processing such as heat transfer and cure kinetics  \cite{johnston1996process},  forming  \cite{boisse2005mesoscopic,hamila2009semi}, stress development  \cite{johnston2001plane, fernlund2003finite}, flow analysis \cite{tan2012multiscale, mohan1999pure, hubert1999two}, and most recently in coupled flow-stress analysis \cite{niaki2017two,niaki2018three,amini20191orthotropic1, amini20192orthotropic2}. However, in situations where repetitive simulations are required---such as optimization, control, real-time monitoring, probabilistic modeling, and uncertainty quantification---the computational cost of the finite element method is prohibitive, as forward simulation involves repeatedly solving large systems of nonlinear equations. \par

To improve the computational efficiency of simulation, the solution of the PDE can instead be directly approximated using a parametrized function. Once trained, the approximate solution can be evaluated for any given input variables with a fixed computational cost. This approach requires two key components. The first is a parametrized function that is flexible enough to approximate the solution of a complex PDE, but simple enough to be efficiently trained and evaluated. Modern \emph{deep neural network} (DNN) models \cite{lecun2015deep}, which are composed of a sequence of linear transformations and component-wise nonlinearities, provide such functions.  DNN models are highly flexible---often having upwards of tens of thousands of parameters---but can still be trained and evaluated efficiently due to recent advances in parallelized hardware, automatic differentiation \cite{bergstra2010theano,baydin2017automatic} and stochastic optimization \cite{ermoliev1988numerical}.\par

The remaining key component to this approach is a loss objective which, when minimized, ensures that the parametrized function is a good approximate solution of the PDE. One option is to evaluate a separate high-fidelity model for a collection of inputs, and then to minimize the difference between these values and predictions for the same inputs. This approach---known as 
\emph{theory-guided machine learning} \cite{karpatne2017theory,wagner2016theory, devries2018deep, liang2018deep, zobeiry2020applications, zobeirytheory}---can be based on a number of models from statistical machine learning such as 
Polynomial Regression \cite{kleijnen2008response}, 
Gaussian process regression (or kriging) \cite{rasmussen2003gaussian},
support vector machines \cite{noble2006support}, 
deep neural networks, and more. 
However, the accuracy of this approach depends heavily on the ability to create a large amount of training data using expensive high-fidelity simulations (e.g., FEM). 

An alternative approach---Physics-Informed Neural Networks (PINNs) \cite{raissi2019physics}---is to train a DNN model to satisfy the PDE directly at a set of training points, known as \emph{sampling} or \emph{collocation} points. By incorporating loss functions based on the PDE, initial condition, and boundary condition residual errors, PINNs recover the solution to the PDE without requiring the use of any other model or simulation. Furthermore, one can include parameters of interest as inputs to the DNN, which enables the use of a PINN as a surrogate model that provides real-time simulation for a range of those parameters.
In contrast to early neural network-based differential equation solvers \cite{meade1994numerical,psichogios1992hybrid,lagaris1998artificial}, PINNs are trained efficiently using the latest advances of DNNs. 
PINNs provide numerous advantages over the finite element method, as well as other surrogate models: it does not require a mesh-based spatial discretization or laborious mesh generation; its derivatives are available once trained; it satisfies the strong form of the differential equations at all training points with known accuracy; it enables integration of data and mathematical models within the same framework \cite{pang2020physics}, and it provides the ability to build surrogates with little or no use of separate high-fidelity simulators. Due to these advantages, many variations of the PINN approach have been successfully used to solve PDEs in a wide range of engineering applications \cite{raissi2019physics,sirignano2018dgm,berg2018unified,anitescu2019artificial,kharazmi2019variational,sun2020surrogate,jagtap2020conservative,nguyen2020deep,goswami2020transfer,haghighat2020deep}. \par

We consider the PINN approach for modelling the thermochemical evolution of a composite-tool system during cure. Developing a PINN for this setting presents major challenges. In particular, the system undergoes two simultaneous coupled processes that exhibit very distinct behaviour: the conduction of heat absorbed from the autoclave and generated by the exothermic composite cure process, and the temperature-dependent curing of the composite \cite{johnston1996process,boyard2016heat}. Capturing both processes simultaneously in a single PINN is nontrivial both for design and training. Past work on the PINN approach in such multiphysics problems \cite{he2020physics} makes use of high-fidelity simulated data---which the PINN method was designed to avoid. Furthermore, the composite-tool system involves a material discontinuity at the composite-tool interface, leading to a discontinuous temperature derivative; standard PINNs \cite{raissi2019physics} produce models with continuous derivatives. While past work has addressed this with domain decomposition \cite{kharazmi2020hp} on the \emph{weak} form of the PDEs, it requires separate basis functions within discretized sub-domains of the geometry leading to a multi-element set-up for learning.

In this paper, we propose novel modifications to the PINN approach designed to address the challenges of modelling composite cure in an autoclave. First, we develop an iterative training procedure for decoupled DNNs to model cure and thermal diffusion that mitigates instability present in standard training methods. In addition, we build explicit discontinuities into the DNN at material boundaries, and enforce known boundary behaviour directly in the PINN loss function. We consider the joint composite-tool system, a realistic resin reaction kinetics model \cite{hubert2001cure}, and a realistic temperature cycle for the autoclave including heat-up, hold, and cool-down regimes---stages which are all important in determining final residual stresses and part geometry after cure. In concurrent work, \citet*{zobeiry2021physics} developed a PINN to solve the heat transfer PDE in a composite-only system (no tool), without resin reaction kinetics, and for a cure cycle excluding the cool-down regime of the processing. As mentioned above, all of these aspects of the problem are crucial to the realistic simulation of the cure process, and are addressed in the present work.

In the remainder of the paper, we present (i) the governing equations for the composite-tool system, (ii) the standard PINN framework, (iii) our proposed modifications to efficiently solve the coupled system of differential equations,  (iv) details of an employed adaptive loss weights scheme, (v) a discussion of how to use transfer learning to reduce training time, (iv) an extension of our method to the surrogate modeling setting with additional input parameters, and (vii) case studies to demonstrate the performance of our approach over a range of composite-tool thicknesses and boundary condition types.

\section{Exothermic Heat Transfer in Curing Composite Materials}\label{sec:formulation}

The exothermic heat transfer in a solid material, i.e. heat transfer with internal heat generation, is governed by the following PDE \cite{johnston1996process, boyard2016heat}
\begin{equation} \label{eq1}
    \frac{\partial}{\partial t} \left(\rho C_p T\right) = 
    \frac{\partial}{\partial x} \left(k_{xx} \frac{\partial T}{\partial x}\right) +
    \frac{\partial}{\partial y} \left(k_{yy} \frac{\partial T}{\partial y}\right) +
    \frac{\partial}{\partial z} \left(k_{zz} \frac{\partial T}{\partial z}\right) +
    \dot{Q},
\end{equation}
where $T$ is the temperature, $C_p$, $k$, and  $\rho$ are the specific heat capacity, conductivity, and density of the solid, respectively, and $\dot{Q}$ is the rate of internal heat generation in the solid. For composite materials with polymeric resin, the internal heat generation is expressed as a function of resin degree of cure $\alpha \in (0,1)$, which is a measure of the conversion achieved during crosslinking reactions of the resin material \cite{johnston1996process}. In particular, the relationship between $\dot{Q}$ and $\alpha$ may be expressed as
\begin{equation} \label{eq2}
    \dot{Q} = \nu_r \rho_r H_r \frac{d \alpha}{d t},
\end{equation}
where subscript $r$ represents resin, $\nu$ is volume fraction, and $H$ is the heat of reaction generated per unit mass of resin during curing. Considering a one-dimensional system with homogeneous material, i.e. $\frac{\partial k}{\partial x} = 0$,  with temperature-independent physical properties, i.e. $\frac{\partial \rho}{\partial t} = \frac{\partial C_p}{\partial t} = 0$, we can rewrite \cref{eq1} as
\begin{equation} \label{eq3}
    \frac{\partial T}{\partial t} = 
    a \frac{\partial^2 T}{\partial x^2}
    + b \frac{d \alpha}{d t} \quad \mathrm{where} \quad a = \frac{k}{\rho C_p} \quad \mathrm{and} \quad b = \frac{\nu_r \rho_r H_r}{\rho C_p}.
\end{equation}
In polymeric composite materials, the evolving degree of cure $\alpha$ is typically governed by an ordinary differential equation that expresses the rate of cure as a function of instantaneous temperature and degree of cure, 
\begin{equation} \label{eq4}
    \frac{d \alpha}{d t} = 
    g(\alpha, T).
\end{equation}
This leads to a coupled system of differential equations (\cref{eq3,,eq4}) for temperature and degree of cure in the space-time domain, $(x, t)$. Solution of this system of differential equations subjected to initial and boundary conditions results in predictions of temperature $T(x,t)$ and degree of cure $\alpha(x,t)$ in the composite material. \par

\begin{figure}[t!]
    \centering
    \includegraphics[width=\textwidth,height=\textheight,keepaspectratio]{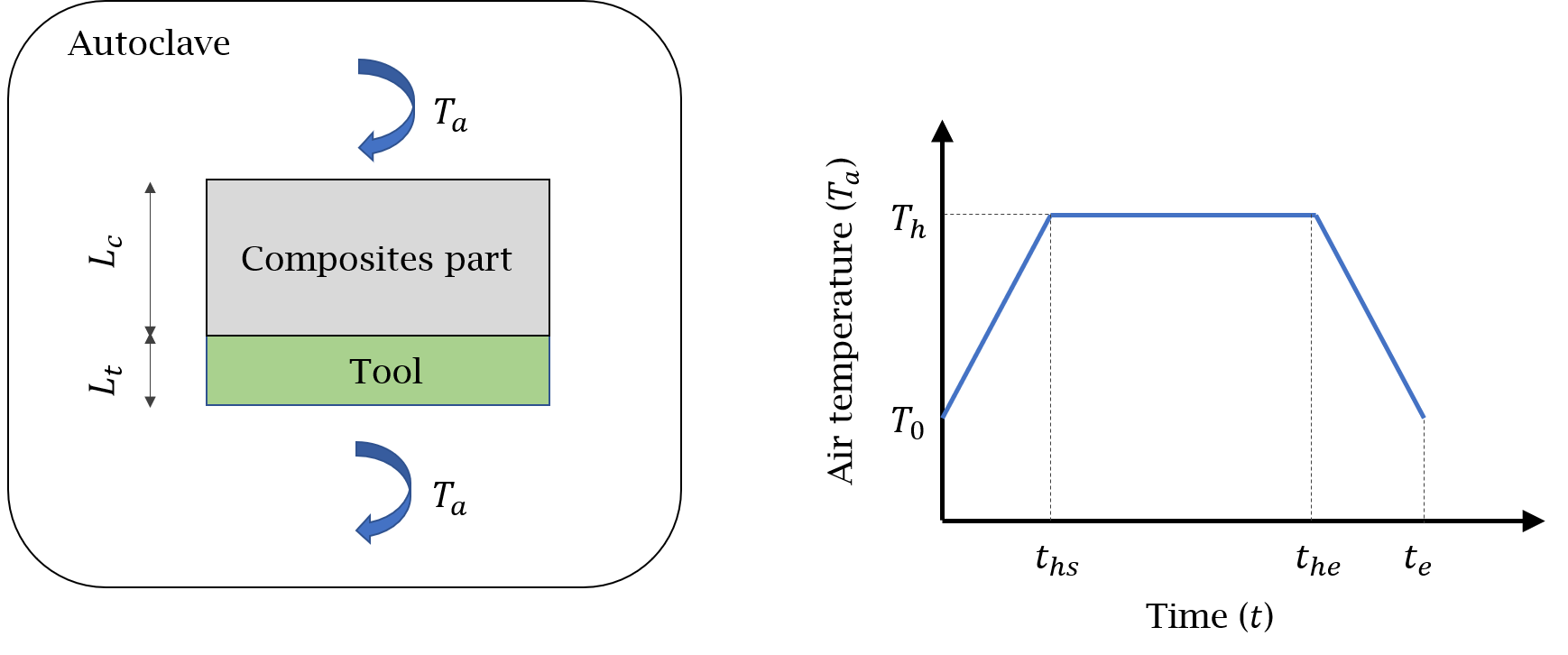}
    \caption{(Left) A schematic of composite-tool material system inside an autoclave; (Right) autoclave air temperature history during the processing time.}
    \label{fig1}
\end{figure}
In this study, we focus on solving the system of differential equations for the composite-tool system shown in \cref{fig1}. Heat transfer in this bi-material system is governed by the differential equations \eqref{eq3} and \eqref{eq4}. However, the material properties are discontinuous at the intersection point of the composite and the tool material, $x=L_t$, i.e.

\begin{equation} \label{eq5}
    \begin{aligned}
        a =
            \begin{cases}
            \begin{aligned}
            a_t \quad &\text{for} \quad \quad 0<x<L_t\\
            a_c \quad &\text{for} \quad \quad L_t<x<L_t+L_c\\
            \end{aligned}
            \end{cases},\\
        b =
            \begin{cases}
            \begin{aligned}
            0 \quad &\text{for} \quad \quad 0<x<L_t\\
            b_c \quad &\text{for} \quad \quad L_t<x<L_t+L_c\\
            \end{aligned}
            \end{cases},
    \end{aligned}
\end{equation}
where the subscripts $c$ and $t$ represent composite and tool materials, respectively. In an autoclave the air temperature, $T_a$, evolves over time during processing; an example is 
shown in \cref{fig1}. Therefore, the boundary conditions at both ends can be specified to be equal to the autoclave air temperature (\emph{prescribed} boundary conditions) via
\begin{equation} \label{eq6}
    \begin{aligned}
        &T|_{x=0} = T_a(t),\\
        &T|_{x=L_t+L_c} = T_a(t).
    \end{aligned}
\end{equation}
However, we may instead capture a more realistic assumption that heat transfers between the air inside the autoclave and the solid materials within it (\emph{convective} boundary conditions) via 
\begin{equation} \label{eq7}
    \begin{aligned}
        &h_t \left( T|_{x=0} - T_a(t)\right) = k_t \frac{\partial T}{\partial x} |_{x=0}, \\
        &h_c \left(T_a(t) - T|_{x=L_t+L_c}\right) = k_c \frac{\partial T}{\partial x} |_{x=L_t+L_c},
    \end{aligned}
\end{equation}
where $h$ is the convective heat transfer coefficient at the interface between the air and the solid. The initial conditions are expressed as
\begin{equation} \label{eq8}
    \begin{aligned}
        &T|_{t=0} = T_0(x),\\
        &\alpha|_{t=0} = \alpha_0(x),
    \end{aligned}
\end{equation}
where $T_0$ is the initial temperature of the system, usually considered constant for the entire geometry, and $\alpha_0$ is the resin initial degree of cure which is considered zero (or a very small number) for uncured resin material over the entire spatial domain.

\section{Physics-Informed Neural Network for Thermochemical Cure Process}\label{sec:PINN}

In this section, we first develop the physics-informed neural network (PINN) formulation for heat transfer in a single material with internal heat generation. Subsequently, we extend the formulation for the more practical composite-tool material system shown in \cref{fig1}.

\subsection{PINN for a single material system}\label{singlematerial}
\begin{figure}[t!]
    \centering
    \includegraphics[width=\textwidth,height=\textheight,keepaspectratio]{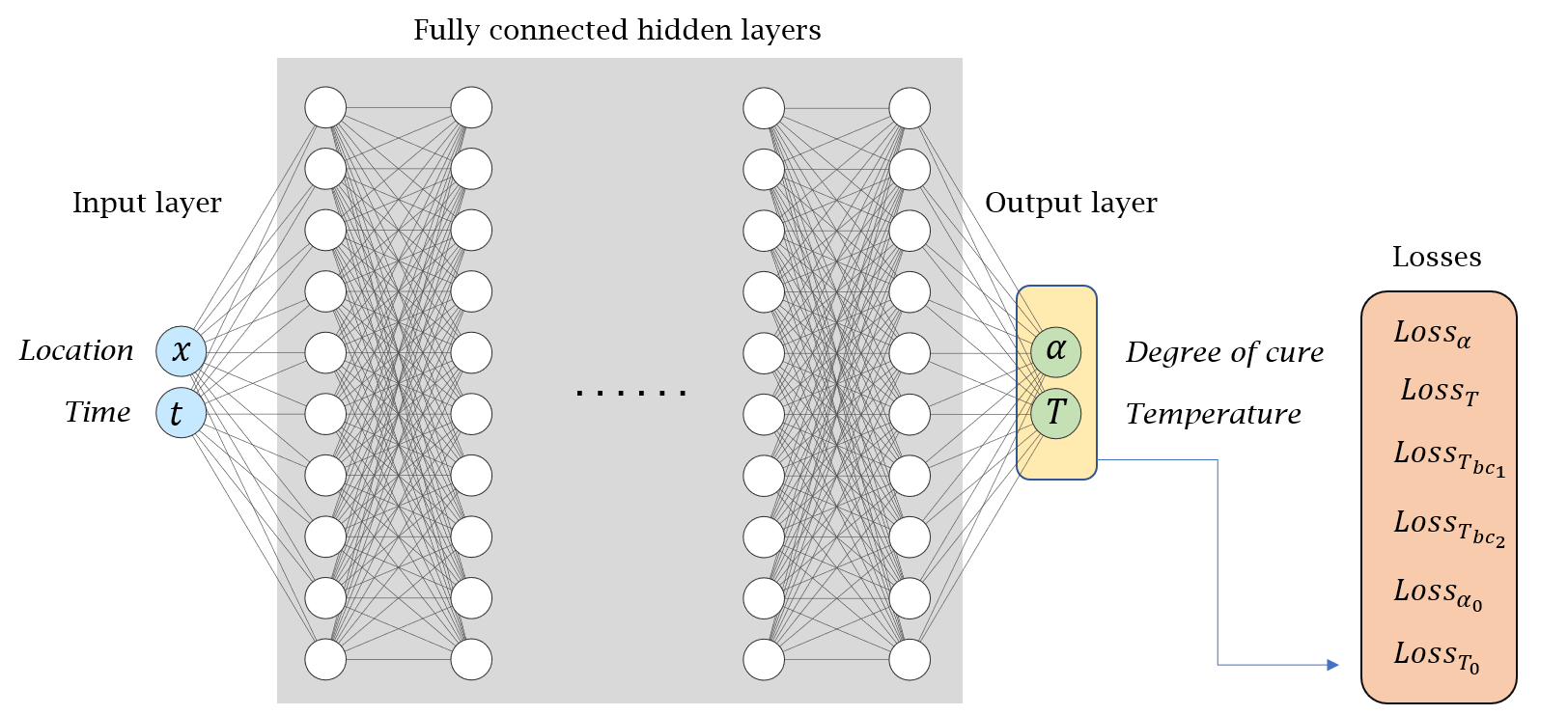}
    \caption{A schematic of a single neural network architecture with space and time input features $(x,t)$ and degree of cure and temperature outputs  $(\alpha, T)$ with a multi-objective optimization of the total loss function consisted of loss terms for governing equations, initial, and boundary conditions.}
    \label{fig2}
\end{figure}

In our heat transfer formulation presented in the previous section, $\boldsymbol{x}:=(x,t)$ is introduced as the set of independent space and time variables $(x,t)$ and $\boldsymbol{y}:=(T,\alpha)$  represents the set of dependent temperature and degree of cure solution variables $(T, \alpha)$. Based on the PINN framework \cite{raissi2019physics}, the solution variable $\boldsymbol{y}$ is approximated using a feed-forward multi-layer neural network 
\begin{equation}\label{eq9}
    \boldsymbol{y}(\boldsymbol{x}) \approx \mathcal{N}(\boldsymbol{x}; \boldsymbol{\theta}),
\end{equation}
where $\mathcal{N}$ is an $\Lambda$-layer neural network with input \emph{features} $\boldsymbol{x} = (x,t)$ and set of all network \emph{parameters} $\boldsymbol{\theta} \in \mathbb{R}^D$, where $D$ is the total number of parameters (equivalent to the total number of \emph{degrees of freedom} in the FEM sense). A schematic of the network for standard PINN is shown in \cref{fig2}. Analytically, this network represents the following compositional function
\begin{equation} \label{eq10}
    \mathcal{N}(\boldsymbol{x}; \boldsymbol{\theta}) = \Sigma^\Lambda \circ \Sigma^{\Lambda-1} \circ \dots \circ \Sigma^{1}(\boldsymbol{x}),
\end{equation}
where
\begin{equation}\label{eq11}
    \Sigma^\lambda({\boldsymbol{z}}^{\lambda-1}) := {\boldsymbol{z}}^\lambda = \sigma^{\lambda}(\mathbf{W}^\lambda \cdot \boldsymbol{z}^{\lambda-1} + \mathbf{b}^\lambda) \quad \text{where} \quad \lambda = 1,...,\Lambda.
\end{equation}
In the above equations, the symbol $\circ$ denotes the composition operation, superscript $\lambda$ is the layer number, $\boldsymbol{z}^{0} := \boldsymbol{x}$ are the inputs $(x,t)$ to the network, $\boldsymbol{z}^{\Lambda} := \boldsymbol{y}$ are the outputs $(T, \alpha)$ of the network, and $\mathbf{W}^\lambda$ and $\mathbf{b}^\lambda$ are the parameters of layer $\lambda$, also known as \emph{weight} matrices and \emph{bias} vectors, all collected in $\boldsymbol{\theta}$.  $\sigma^\lambda$ are so-called activation functions which introduce nonlinearity into the neural network \cite{sibi2013analysis}. Typically a single function---such as the \emph{hyperbolic-tangent} function---is selected for all layers except for the last layer, $\sigma^\Lambda$, which is selected based on the output variables, here $T$ and $\alpha$.  \par

If the activation function $\sigma$ is continuous and differentiable, the neural network in \cref{eq10} will have these properties over the $\boldsymbol{x}$ domain as well. In PINNs, an error or \emph{loss} function is  defined using the processed outputs of the network and their derivatives based on the equations governing the physics of the problem. In particular for our problem, the total loss of the network, $\mathcal{L}_\mathcal{N}$, consists of the sum of loss terms for 
the PDE ($\mathcal{L}_\alpha$ and $\mathcal{L}_T$),
the initial conditions ($\mathcal{L}_{\alpha_0}$ and  $\mathcal{L}_{T_0}$),
and the boundary conditions ($\mathcal{L}_{T_{bc_1}}$ and $\mathcal{L}_{T_{bc_2}}$),
\begin{equation}\label{eq125}
    \mathcal{L}_\mathcal{N} = \mathcal{L}_\alpha + \mathcal{L}_T + \mathcal{L}_{\alpha_0} +\mathcal{L}_{T_0} + \mathcal{L}_{T_{bc_1}} + \mathcal{L}_{T_{bc_2}},
\end{equation}
where each term is evaluated on a set of training or \emph{collocation} points in the domain denoted as $\mathbf{X}$.  
The PDE losses associated with \cref{eq3,,eq4}, in the form of mean squared error terms, are expressed as
\begin{equation} \label{eq13}
    \begin{aligned}
        \mathcal{L}_\alpha &=  
        \frac{1}{n}\sum_{i=1}^{n} \left(\frac{d \alpha}{d t}|_{\boldsymbol{x}_i} -
        g(\alpha, T)|_{\boldsymbol{x}_i} \right)^2,\\
        \mathcal{L}_T &= \frac{1}{n}\sum_{i=1}^{n} \left(\frac{\partial T}{\partial t}|_{\boldsymbol{x}_i} -
        a \frac{\partial^2 T}{\partial x^2}|_{\boldsymbol{x}_i}
        - b \frac{d \alpha}{d t}|_{\boldsymbol{x}_i} \right)^2,\\
    \end{aligned}
\end{equation}
 where $n$ is the total number of collocation points and $\boldsymbol{x}_i \in \mathbf{X}$. For the case of prescribed temperature boundary conditions, the losses associated with initial and boundary conditions \cref{eq6,,eq8} are defined as 
\begin{equation} \label{eq14}
    \begin{aligned}
        \mathcal{L}_{\alpha_{0}} &=  
        \frac{1}{n}\sum_{i=1}^{n} \left(\alpha|_{t=0} -\alpha_0(x) \right)^2,\\
        \mathcal{L}_{T_{0}} &=  
        \frac{1}{n}\sum_{i=1}^{n} \left(T|_{t=0} - T_0(x)\right)^2,\\
        \mathcal{L}_{T_{bc_1}} &=  
        \frac{1}{n}\sum_{i=1}^{n} \left(T|_{x=0} - T_a(t) \right)^2,\\
        \mathcal{L}_{T_{bc_2}} &=  
        \frac{1}{n}\sum_{i=1}^{n} \left(T|_{x=L_t+L_c} - T_a(t)\right)^2.\\
    \end{aligned}
\end{equation}
For the convective heat transfer boundary condition expressed in \cref{eq7}, the last two terms in the above equation, i.e. $\mathcal{L}_{T_{bc_1}}$ and $\mathcal{L}_{T_{bc_2}}$, are replaced with 
\begin{equation} \label{eq15}
    \begin{aligned}
        \mathcal{L}_{T_{bc_1}} &=  
        \frac{1}{n}\sum_{i=1}^{n} \left(h_t \left( T|_{x=0} - T_a(t)\right) - k_t \frac{\partial T}{\partial x} |_{x=0}\right)^2,\\
        \mathcal{L}_{T_{bc_2}} &=  
        \frac{1}{n}\sum_{i=1}^{n} \left(h_c \left(T_a(t) - T|_{x=L_t+L_c}\right) - k_c \frac{\partial T}{\partial x}|_{x=L_t+L_c}\right)^2.\\
    \end{aligned}
\end{equation}
The network is then trained by minimizing the total loss for the set of network parameters $\boldsymbol{\theta}$
at the collocation points $\mathbf{X}$, \par
\begin{equation}\label{eq16}
    \boldsymbol{\theta}^* = \argmin_{\boldsymbol{\theta} \in \mathbb{R}^D} \mathcal{L}_\mathcal{N} (\mathbf{X}; ~\boldsymbol{\theta}).
\end{equation}
It is worth noting that the loss terms include derivatives of the DNN function approximations of $T$ and $\alpha$; it is possible to compute these exactly using automatic differentiation \cite{baydin2017automatic} at any point in the domain without the need for manual computation.
\begin{figure}[t!]
    \centering
    \includegraphics[width=\textwidth,height=\textheight,keepaspectratio]{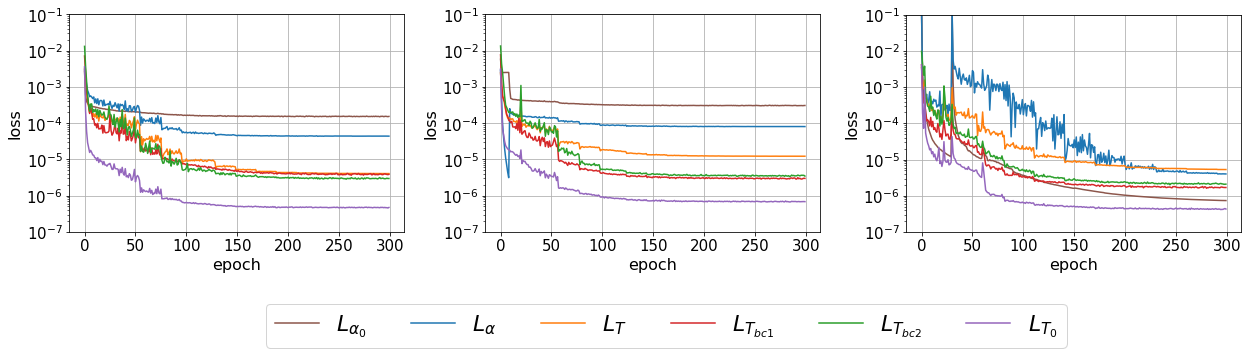}
    \caption{Evolution of different loss terms in the training of the network. (Left) A joint network architecture, 5 layers and 42 nodes (7436 parameters in total); (Middle) a disjoint network architecture with simultaneous training, 5 layers and 30 nodes for each network (7682 parameters in total); (Right) a disjoint network architecture with the proposed sequential training procedure, 5 layers and 30 nodes for each network (7682 parameters in total).}
    \label{fig0}
\end{figure}

\begin{figure}[t!]
    \centering
    \includegraphics[width=\textwidth,height=\textheight,keepaspectratio]{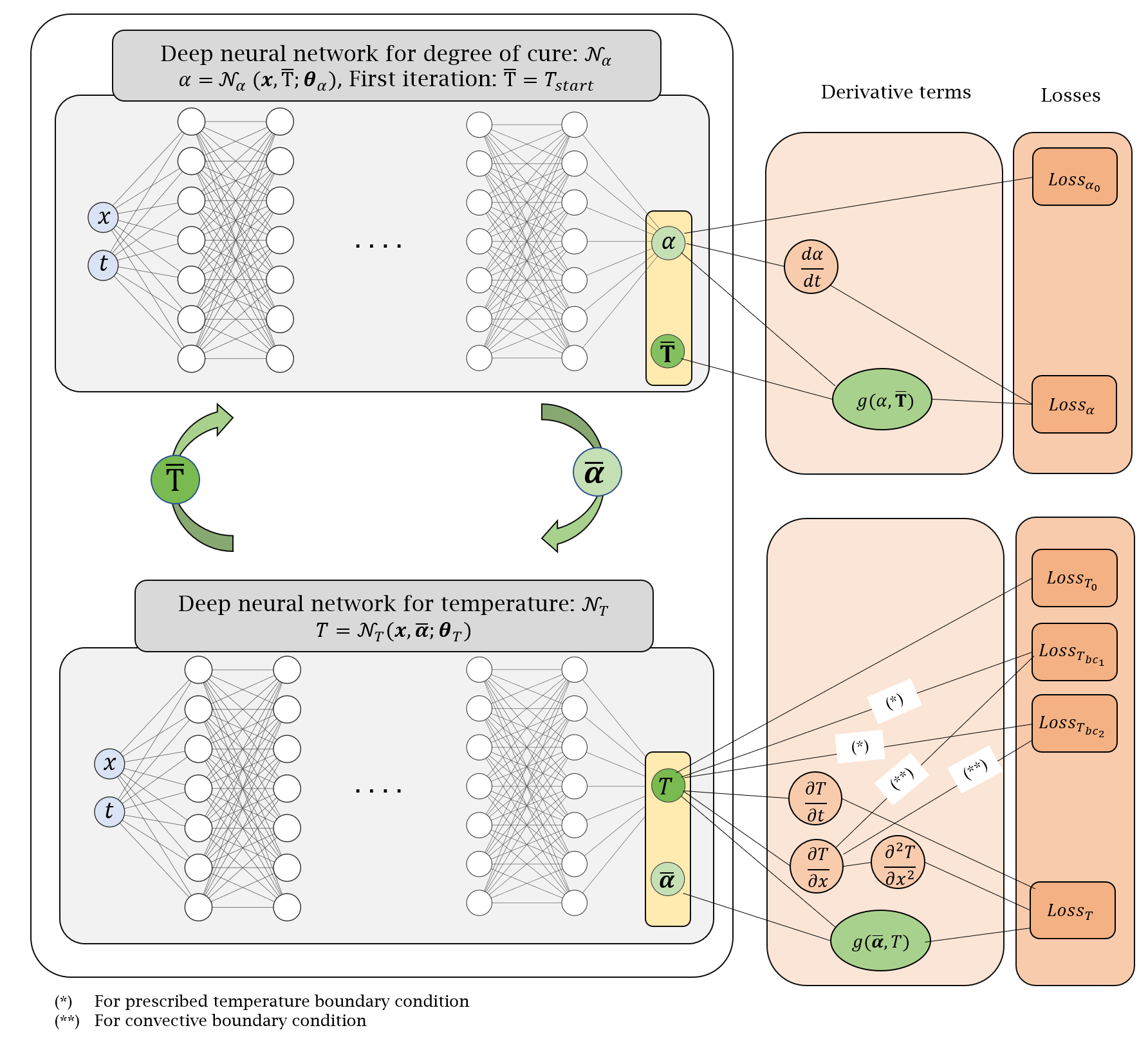}
    \caption{A schematic of network architecture for a single material system. Two separate networks are constructed for $T$ and $\alpha$ with sequential training first on $\alpha$, and then on $T$, until convergence.}
    \label{fig3}
\end{figure}

However, given the coupling of $\mathcal{L}_\alpha$ and $\mathcal{L}_T$ loss terms in \cref{eq13}, training a single network for both $\alpha$ and $T$ outputs is not straightforward.
For example, \cref{fig0} (left) shows that standard minimization algorithms fail to reduce the losses associated with the $\alpha$ output to an acceptable level.
Here we take inspiration from classical solution methods (e.g., FEM), where each output is approximated independently without sharing degrees of freedom. Rather than a single network, we train two separate networks---one for each output variable, $\alpha$ and $T$---in a manner that resembles past disjoint PINNs \cite{haghighat2020deep, zhang2019quantifying, he2020physics, sahli2020physics},
\begin{equation} \label{eq17}
    \begin{aligned}
        &\alpha(\boldsymbol{x}) \approx \mathcal{N}_{\alpha}(\boldsymbol{x};~\boldsymbol{\theta}_{\alpha}), \\
        &T(\boldsymbol{x}) \approx \mathcal{N}_{T}(\boldsymbol{x};~\boldsymbol{\theta}_{T}),
    \end{aligned}
\end{equation}
where $\boldsymbol{\theta}_{\alpha} \in \mathbb{R}^{D_\alpha}$ and $\boldsymbol{\theta}_{T} \in \mathbb{R}^{D_T}$ are the parameters for the individual networks. 
In addition, even with separate networks, simultaneous training of both networks also tends to fail as demonstrated in \cref{fig0} (middle). 
To resolve this, we propose the use of sequential training for $\alpha$ and $T$. In particular, we first optimize the network parameters of $\mathcal{N}_\alpha$
with those of $\mathcal{N}_T$ fixed, then vice versa, and the process is iterated until convergence is achieved. 
In this sequential approach, the total loss in \cref{eq125} is then decoupled as
\begin{equation}\label{eq175}
\begin{aligned}
    \mathcal{L}_{\mathcal{N}_\alpha} &= \mathcal{L}_\alpha + \mathcal{L}_{\alpha_0}, \\
    \mathcal{L}_{\mathcal{N}_T} &= \mathcal{L}_T + \mathcal{L}_{T_0} + \mathcal{L}_{T_{bc_1}} + \mathcal{L}_{T_{bc_2}}.
\end{aligned}
\end{equation}
The optimization problem for the $j$th iteration of sequential training, with initial network parameters ${\boldsymbol{\theta}^*_\alpha}^{j}$ and ${\boldsymbol{\theta}^*_T}^{j}$, is defined as
\begin{equation}\label{eq18}
\begin{aligned}
    {\boldsymbol{\theta}^*_\alpha}^{j+1} &= \argmin_{\boldsymbol{\theta}_{\alpha} \in \mathbb{R}^{D_\alpha}} \mathcal{L}_{\mathcal{N}_\alpha} (\mathbf{X}, \bar{\mathbf{T}}^{j}; {\boldsymbol{\theta}}_\alpha), \\
    {\boldsymbol{\theta}^*_T}^{j+1} &= \argmin_{\boldsymbol{\theta}_{T} \in \mathbb{R}^{D_T}} \mathcal{L}_{\mathcal{N}_T} (\mathbf{X}, \bar{\boldsymbol{\alpha}}^{j+1}; {\boldsymbol{\theta}_T}),  
\end{aligned}
\end{equation}
with $\bar{\boldsymbol{\alpha}}^{j} = \mathcal{N}_{\alpha}(\mathbf{X};{\boldsymbol{\theta}^*_T}^{j})$ and $\bar{\mathbf{T}}^j = \mathcal{N}_{T}(\mathbf{X}; {\boldsymbol{\theta}^*_\alpha}^{j})$ as the sequentially approximated values of solution variables in \cref{eq17} on the training points $\mathbf{X}$. A schematic architecture of the proposed network and the sequential training procedure is shown in \cref{fig3}.
The convergence of the various losses in the proposed PINN using the proposed sequential training procedure and decoupled networks for $\alpha$ and $T$ also is illustrated in \cref{fig0} (right). In this figure, the $\alpha$-related losses are trained for 30 epochs, then the $T$-related losses for 30 epochs, and this is repeated for 10 iterations. The sequential procedure successfully reduces all loss terms to a level below $10^{-5}$.

\subsection{PINN for a bi-material system}\label{bimaterial}
By construction, the feed-forward multi-layer neural network \eqref{eq9} (and both networks in \cref{eq17}) with a differentiable activation function, e.g. hyperbolic-tangent, represents a continuous and differentiable function in space and time for any given $\boldsymbol{x}$ in the domain. However, for a bi-material system where the material properties are discontinuous at the interface, the derivative of solution variables is discontinuous at that point. To improve the accuracy of the neural network approximations for problems where we expect such discontinuous solutions at the bi-material interface $x=x_I$, where $x_I=L_t$ in \cref{fig1}, we propose the following construction for both neural networks $\mathcal{N}_\alpha$ and $\mathcal{N}_T$:
\begin{equation} \label{eq19}
    \mathcal{N}(\boldsymbol{x}; \boldsymbol{\theta}) = H(x_I - x)\mathcal{N}^-(\boldsymbol{x}; ~\boldsymbol{\theta}^-) + H(x - x_I)\mathcal{N}^+(\boldsymbol{x}; ~\boldsymbol{\theta}^+), 
\end{equation}
where $H$ is the heaviside step function defined as
\begin{equation} \label{eq20}
    H(x) = \begin{cases}
        1 & \text{ if } x\ge0 \\ 
        0 & \text{ if } x<0.
       \end{cases}
\end{equation}
This construction of a neural network using two components $\mathcal{N}^-$ and $\mathcal{N}^+$ with distinct parameters $\boldsymbol{\theta}^-$ and $\boldsymbol{\theta}^+$, implies that the space of possible solutions for $\boldsymbol{y} \approx \mathcal{N}(\boldsymbol{x}; \boldsymbol{\theta})$ and its spatial derivative can be discontinuous at $x = x_I$, i.e.

\begin{equation} \label{eq21}
\begin{aligned}
        \boldsymbol{y}|_{x^+}  - \boldsymbol{y}|_{x^-} &= \mathcal{N}^+(\boldsymbol{x}; ~\boldsymbol{\theta}^+) - \mathcal{N}^-(\boldsymbol{x}; ~\boldsymbol{\theta}^-), \\
        \frac{\partial {\boldsymbol{y}}}{\partial x}|_{x^+} -
    \frac{\partial {\boldsymbol{y}}}{\partial x}|_{x^-} &= \frac{\partial\mathcal{N}^+(\boldsymbol{x}; ~\boldsymbol{\theta}^+)}{\partial {x}} - \frac{\partial \mathcal{N}^-(\boldsymbol{x}; ~\boldsymbol{\theta}^-)}{\partial {x}}.
\end{aligned}
\end{equation}
Therefore, the functional form \eqref{eq19} can approximate discontinuous solution at $x=x_I$. 

\begin{figure}[t!]
    \centering
    \includegraphics[width=\textwidth,height=\textheight,keepaspectratio]{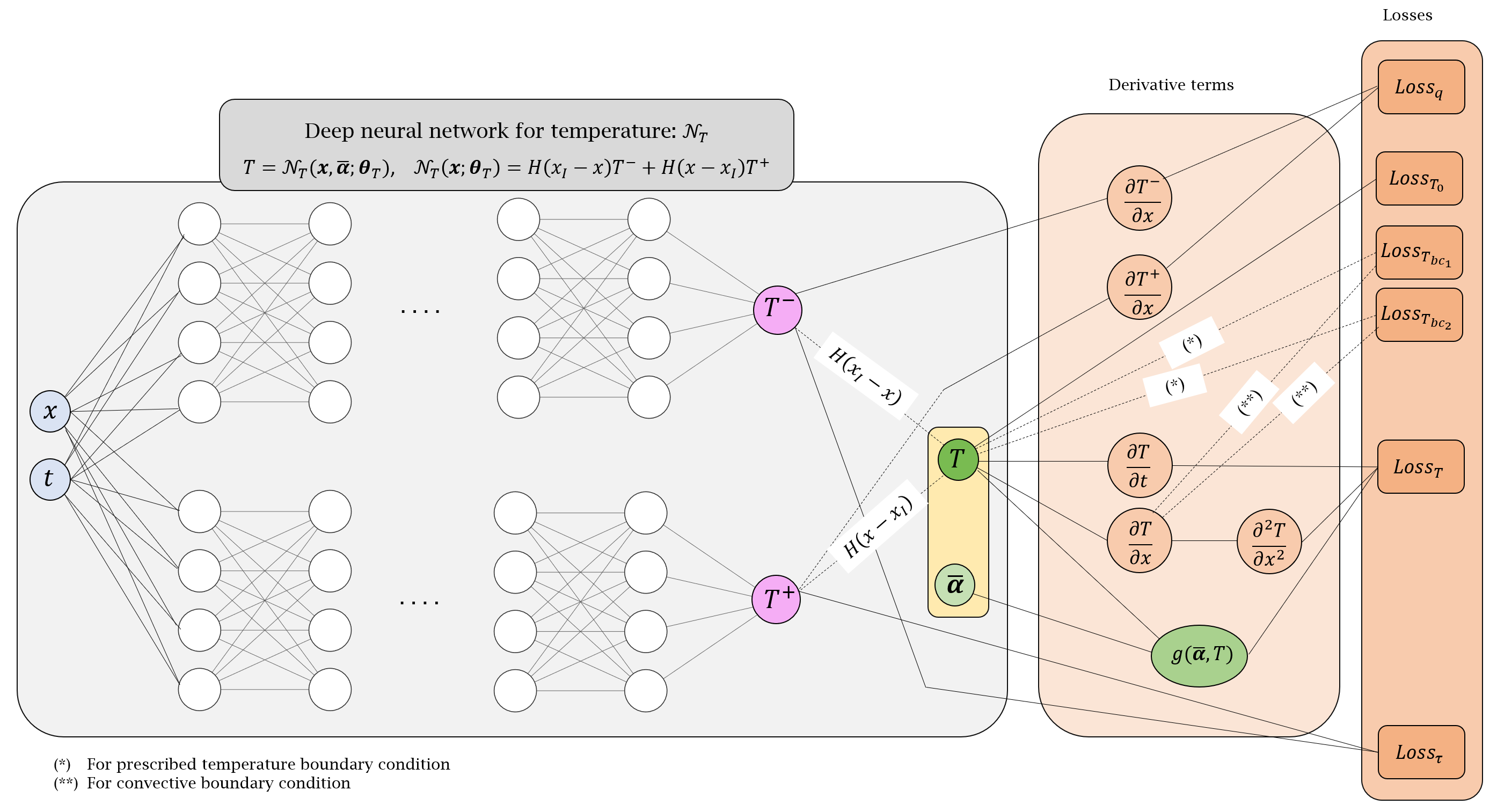}
    \caption{A schematic of a discontinuous network architecture utilized for the bi-material problem. Two separate networks $T^-$ and $T^+$ are combined with Heaviside step function $H$ to construct the solution space for $T$.}
    \label{fig4}
\end{figure}

For the bi-material composite-tool system, the degree of cure is only applicable to the composite material. The temperature distribution, however, is defined for both the tool and the composite part. Following the previous discussions on the choice of networks, the solution space in \cref{eq17} is then defined as
\begin{equation} \label{eq22}
    \begin{aligned}
        &\alpha(\boldsymbol{x}) \approx H(x - x_I)\mathcal{N}_{\alpha}^+(\boldsymbol{x};~\boldsymbol{\theta}_{\alpha}^+) , \\
        &T(\boldsymbol{x}) \approx H(x_I - x) \mathcal{N}^-_{T}(\boldsymbol{x};~\boldsymbol{\theta}_{T}^-) + H(x - x_I) \mathcal{N}^+_{T}(\boldsymbol{x};~\boldsymbol{\theta}_{T}^+),
    \end{aligned}
\end{equation}
which is schematically shown in \cref{fig4} where the $T^{-}$ and $T^{+}$ nodal values are
\begin{equation} \label{eq70}
    \begin{aligned}
        &T^{-} = \mathcal{N}^-_{T}(\boldsymbol{x};~\boldsymbol{\theta}_{T}^-),\\
        &T^{+} = \mathcal{N}^+_{T}(\boldsymbol{x};~\boldsymbol{\theta}_{T}^+).
    \end{aligned}
\end{equation}

At the interface point between the tool and the composite, i.e. at $x=x_I$, the temperature field should remain continuous. However, as shown in \cref{eq21}, the temperature from the neural network is allowed to be discontinuous. Therefore, the optimization should satisfy $T|_{x=x_I^+}-T|_{x=x_I^-}=0$ in order to impose the continuity condition for the temperature. Additionally, although the spatial derivative $\frac{\partial T}{\partial x}$ may be discontinuous, the heat conservation implies that the heat flux across the interface should also be continuous, i.e., $k^+ \frac{\partial {T^+}}{\partial x}|_{x=x_I} - k^- \frac{\partial {T^-}}{\partial x}|_{x=x_I} = 0$.
The loss terms for these two conditions are expressed as 
\begin{equation} \label{eq23}
    \begin{aligned}
        \mathcal{L}_\tau &= \sum_{i=1}^{n} \left(T^+|_{x=x_I} - T^-|_{x=x_I}\right)^2, \\
        \mathcal{L}_q &= \sum_{i=1}^{n} \left(k^+ \frac{\partial T^+}{\partial x}|_{x=x_I} - k^- \frac{\partial T^-}{\partial x}|_{x=x_I} \right)^2,
    \end{aligned}
\end{equation}
where $k^-$ and $k^+$ are $k_t$ and $k_c$, respectively, and $x_I=L_t$ in the composite-tool setup in \cref{fig1}. Based on the cases we studied, this choice of network is contributing strongly to accurate modeling of the bi-material composite-tool system. \par

\subsection{Network layers}\label{PINNdetails}
For training of PINN described above, it is crucial to use activation functions
tailored to the underlying differential equations and the physical output
quantities. Using physics based rationale, an activation function with positive
output parameter and non-zero derivative (which is important in physics based
neural networks) is an obvious choice for the last layer of $\mathcal{N}_T$.
Therefore, in order to result in a positive absolute temperature $T>0$, the
\emph{Softplus} function is used which is defined as
\begin{equation} \label{eq24}
    \text{Softplus}(x) = \text{ln}(1+e^x).
\end{equation}
For the degree of cure, which is defined in $ \alpha \in (0, 1)$, the \emph{Sigmoid} function is a natural choice for the last layer of $\mathcal{N}_\alpha$ where its following modified version is used in those cases where curing of the composite starts from a nonzero value, $\alpha_0$,
\begin{equation} \label{eq26}
    \text{Sigmoid} ' (x) = \frac{1-\alpha_0}{1+e^{-x}} + \alpha_0.
\end{equation}
For all hidden layers, hyperbolic-tangent function is used which is a preferable activation function due to its smoothness and non-zero derivative. \par

\subsection{Adaptive loss weights}\label{adaptiveloss}
As discussed by others \cite{raissi2019physics}, training the multi-objective total loss functions, e.g. $\mathcal{L}_{\mathcal{N}_\alpha}$ and $\mathcal{L}_{\mathcal{N}_T}$ in \cref{eq175}, poses challenges for optimization techniques. Particularly for boundary value problems, the unique solution to the governing equations is achieved when initial and boundary conditions are imposed strongly. Within the context of multi-objective optimization, a weighted-sum scheme is commonly employed to improve the optimization algorithms \cite{marler2010weighted}. In practice, the loss weights are often tuned manually in a very tedious trial-and-error procedure based on the convergence analysis of the optimization process. Within the context of PINNs, \citet{wang2020understanding} showed that inaccuracy in optimization is particularly affected by \emph{unbalanced gradients} and proposed an adaptive loss weight algorithm based on normalizing the gradient of individual terms in the loss function in order to reduce the \emph{stiffness of the gradient flow dynamics}. Therefore, the total loss terms $\mathcal{L}_{\mathcal{N}_\alpha}$ and $\mathcal{L}_{\mathcal{N}_T}$ in \cref{eq175} are modified as
\begin{equation} \label{eq265}
    \begin{aligned}
        \mathcal{L}_{\mathcal{N}_\alpha} &=  \mathcal{L}_\alpha + \omega_{\alpha_0} \mathcal{L}_{\alpha_0}, \\
        \mathcal{L}_{\mathcal{N}_T} &=  \mathcal{L}_T + \omega_{T_0} \mathcal{L}_{T_0} + \omega_{T_{bc_1}} \mathcal{L}_{T_{bc_1}} + \omega_{T_{bc_2}} \mathcal{L}_{T_{bc_2}},
    \end{aligned}
\end{equation}
where $\omega$ denotes the loss weight corresponding to each loss term. The updated scaling weight $\omega^{e+1}$ for each loss term is evaluated based on the parameter values $\boldsymbol{\theta}^e$ at epoch $e$ as
\begin{equation} \label{eq266}
    \omega^{e+1} = \beta \omega^{e} + (1-\beta) \hat{\omega}^{e+1},
\end{equation}
where
\begin{equation} \label{eq267}
    \begin{aligned}
        \hat{\omega}_{\alpha_0}^{e+1} &= \frac{1}{\omega_{\alpha_0}^{e}}\frac{\text{max} \left(|\nabla_{\theta} \mathcal{L}_\alpha(\boldsymbol{\theta}^e) |\right)}
        {\text{mean}\left(|\nabla_\theta \mathcal{L}_{\alpha_0}(\boldsymbol{\theta}^e)|\right)} \quad &\mathrm{for}& \quad \mathcal{N}_\alpha, \\
        \hat{\omega}_i^{e+1} &= \frac{1}{\omega_i^{e}}\frac{\text{max} \left(|\nabla_{\theta} \mathcal{L}_T(\boldsymbol{\theta}^e) |\right)}
        {\text{mean}\left(|\nabla_\theta \mathcal{L}_i(\boldsymbol{\theta}^e)|\right)} \quad &\mathrm{for}& \quad \mathcal{N}_T, \quad \mathrm{where} \quad i \in \{T_0, T_{bc_1}, T_{bc_2} \}.
    \end{aligned}
\end{equation}
In the above equation, $\nabla_\theta$ shows gradient of the loss with respect to the parameter of the network and $\beta$ is taken as 0.9. Note that relations \cref{eq266,eq267} are based on the actual implementation shared by the authors, that are slightly different from the reported relations by  \citet{wang2020understanding}. The loss weight updates could be performed at every epoch or at a frequency specified by the user \cite{wang2020understanding}. Such adaptive method for loss weights enhances robustness of the method and leads to improvement in accuracy of the model predictions.

\subsection{Transfer learning}\label{sec:trasnferlearning}
Training PINN models can be computationally expensive due to the use of first-order
optimization methods. However, in cases where one needs the solution to the differential
equations for a number of related problem parameter settings---e.g., when one is exploring
how perturbing the material properties, geometry, or boundary conditions 
affects the solution---one can achieve a significant reduction
in training time by using \emph{transfer learning}. In particular, after an initial (expensive)
training of the model for a particular parameter setting, one can use the
weight matrices and bias vectors of the neural
network ($\mathbf{W}^\lambda$ and $\mathbf{b}^\lambda$ in \cref{eq10} for
all layers) as initial values of the weights and biases for each similar problem. Using this warm-start initialization strategy, the network
converges much faster for each new problem, resulting in a significant reduction
in overall training time.

\subsection{Surrogate modeling}\label{sec:surrogatemodeling}
In the setting of \emph{surrogate modeling}, we require the solution of the differential
equations (and possibly its gradients) not only for different spatial and temporal domain points, but also for different
values of problem parameters (e.g., material properties, boundary conditions, etc).
The long running time and lack of analytical gradients of traditional high-fidelity 
simulation models, such as FEM, is a major obstacle 
to their use as surrogate models \cite{forrester2008engineering}.
Although a growing body of recent work specifically addresses the surrogate modeling setting 
\cite{queipo2005surrogate, jiang2020surrogate}, these methods typically involve training machine learning models
with data from high-fidelity simulators such as FEM; as large amounts of data are typically required,
this creates a significant and potentially intractable up-front computational cost.

\begin{figure}[ht!]
    \centering
    \includegraphics[width=\textwidth,height=10cm,keepaspectratio]{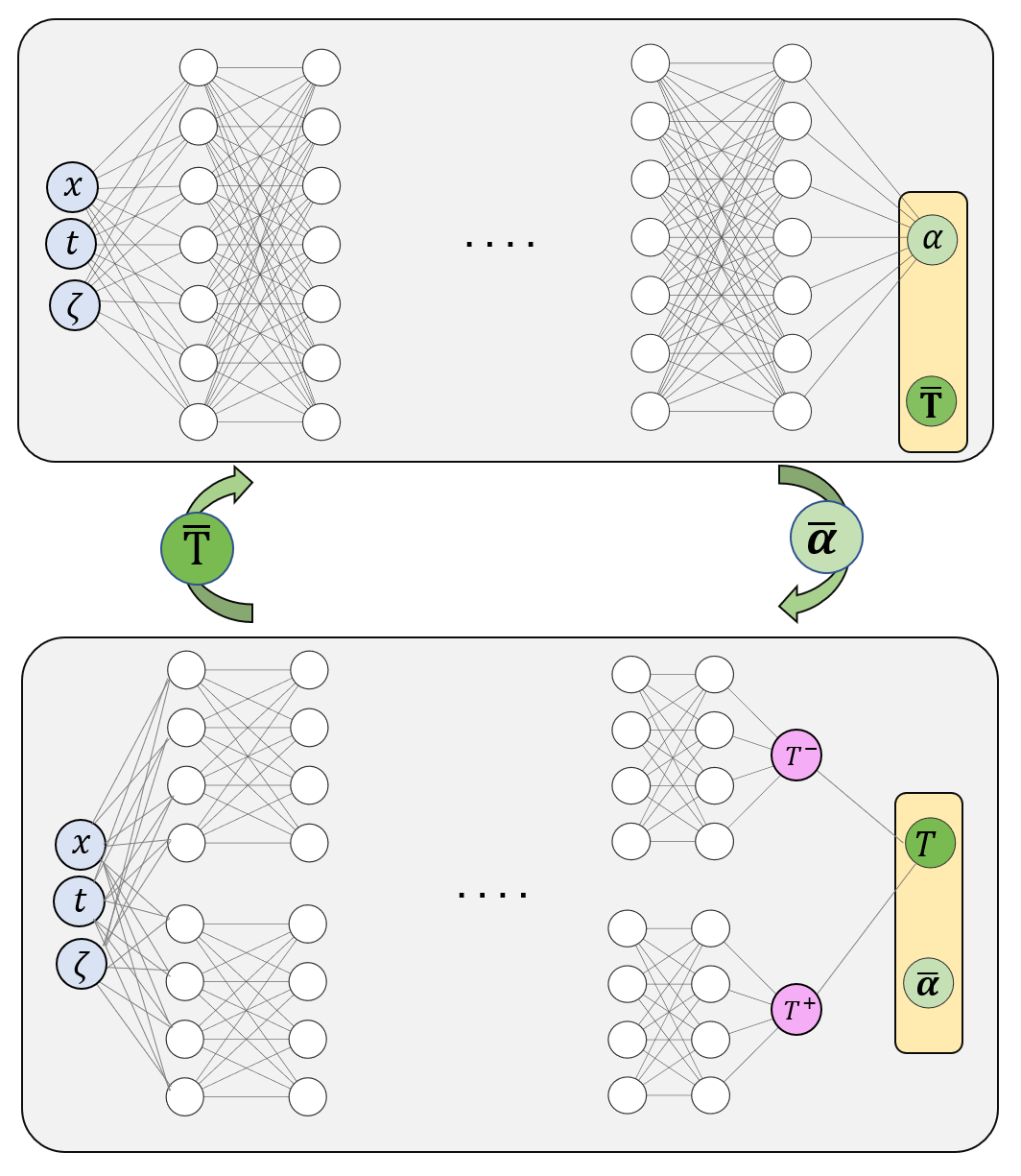}
    \caption{A schematic of network architecture used for constructing a PINN-based surrogate model with one extra input variable, shown as $\zeta$. After training, the model predicts temperature and degree of cure at any new input variables on the spatio-temporal domain.}
    \label{fig15}
\end{figure}

PINNs, on the other hand, offer a natural construction of surrogates that
does not rely on large amounts of pre-generated data \cite{zhu2019physics}. 
In particular, one can simply extend the DNN to include problem parameters
as inputs, and duplicate the original collection of training loss functions for a set
of values within the desired range of those problem parameters.
This methodology does not require any additional data from a separate
high-fidelity simulation; but one can naturally include any data that is 
available with additional losses that directly penalize the PINN output
at the corresponding input values \cite{haghighat2020deep}. For example, 
\cref{fig15} shows the temperature and
degree of cure PINN augmented with an additional parameter $\zeta$, thereby creating
a surrogate composite-tool system model. Once trained, the model can be
used to predict response variables in almost real-time.

\section{Case studies}\label{sec:CaseStudies}

To check the performance of the proposed PINN, the exothermic heat  transfer in a AS4/8552 composite part on an Invar tool is modelled. The resin reaction in 8552 epoxy follows a cure kinetics differential equation presented in \cite{hubert2001cure} as
\begin{equation} \label{eq225}
    \begin{aligned}
        &\frac{ \mathrm{d} \alpha}{ \mathrm{d}t} = \frac{A \; \exp \left(\frac{-\Delta E}{R \; T}\right)}{\exp\left(C\left(\alpha-C_T T-C_0\right)\right)} \; \alpha^M (1-\alpha)^N,
    \end{aligned}
\end{equation}
where the right hand side of the equation represents $g(\alpha, T)$. All parameters in \cref{eq225} are listed in Table \ref{table2} along with material properties for AS4 fibre, 8552 epoxy resin, and Invar tool in \cref{table3}. The composite properties in the transverse direction, i.e. the through thickness, are obtained from those of the fibre and resin  \cite{hubert2001cure, twardowski1993curing} as
\begin{equation} \label{eq234}
    \begin{aligned}
        \rho_c &= \rho_r \nu_r + \rho_f \nu_f, \\
        C_{p_c} &= C_{p_r} \nu_r + C_{p_f} \nu_f, \\
        k_c &= k_r \left( \left(1-2\Omega \right) + \frac{1}{\Omega}  \left(\pi - \frac{4}{\Upsilon} \tan^{-1} \left( \frac{\Upsilon}{ 1 + \Omega \Gamma} \right) \right) \right),
    \end{aligned}
\end{equation}
where
\begin{equation} \label{eq235}
    \begin{aligned}
    \Gamma &= 2 \left(\frac{k_r}{k_f} - 1\right), \\
    \Omega &= \sqrt{\frac{\nu_f}{\pi}}, \\
    \Upsilon &= \sqrt{1 - \Gamma^2 \Omega^2}. 
    \end{aligned}
\end{equation}

\begin{table}[t!]
\centering
    \begin{tabular}{ |l|c| }
        \hline
            \textbf{Parameter} & \textbf{Value}\\
        \hline
            $A$ ($1/\text{s}$) & $1.528 \times 10^5$ \\
            $\Delta E$ ($\text{J}/\text{mol}$) & $6.650 \times 10^4$ \\
            $M$ & 0.8129 \\
            $N$ & 2.7360 \\
            $C$ & 43.09 \\
            $C_0$ & -1.6840 \\
            $C_T$ ($1/\text{K}$)& $5.475 \times 10^{-3}$ \\
            $R$ ($\text{J}/ \text{mol K}$)& 8.314 \\
         \hline
    \end{tabular} 
    \caption{Parameters for resin 8552 cure kinetics model in \cref{eq225} \cite{hubert2001cure}.}
    \label{table2}
\end{table}

\begin{table}[t!]
\centering
    \begin{tabular}{ |l|c|c|c|}
         \hline
             \multirow{2}{6cm}{\textbf{Material parameter}}  & \textbf{AS4 fibre} & \textbf{8552 resin \cite{hubert2001cure}} & \textbf{Invar tool} \\
             & ($f$) & ($r$) &  ($t$) \\
         \hline
             Volume fraction, $\nu$ & 0.574 & 0.426 &  -\\
             Density, $\rho$ ($\text{kg}/\text{m}^3$)   & 1790    & 1300  & 8150\\
             Conductivity, $k$ ($\text{W}/\text{m K}$) & 3.960  & 0.212   & 13.0\\
             Specific heat capacity, $C_p$ ($\text{J}/\text{kg K}$) & 914.0 & 1304.2 & 510.0\\
         \hline
    \end{tabular}
    \caption{Thermal material properties of fibre, resin, and tool.}
    \label{table3}
\end{table}
\par

Results are presented for four case studies with different boundary conditions and different thicknesses of the composite and the tool, as listed in \cref{table4}. For cases 1 and 2, the thickness of composite is $L_c= \SI{30}{\mm}$ which sits on a $L_t = \SI{20}{\mm}$ tool, while for cases 3 and 4, they are taken as $L_c= \SI{300}{\mm}$  and $L_t= \SI{200}{\mm}$. Cases 1 and 3 are subjected to prescribed temperature at the boundary while cases 2 and 4 are subjected to convective boundary conditions. For all cases, the autoclave air temperature follows a heat up-hold-cool down cycle as shown in Fig. \ref{fig1} with $T_0=\SI{293}{\K} (\SI{20}{\celsius})$, $T_h= \SI{453}{\K} (\SI{180}{\celsius})$, $t_{hs}= \SI{52}{\min}$, $t_{he}= \SI{172}{\min}$, and $t_{e}= \SI{222}{\min}$. The initial temperature of both composite and tool are prescribed as $T_0=\SI{293}{\K}$. The training is conducted on a uniform grid with 500 points on $0<x<L_t+L_c$ and 1,000 points on $0<t<t_e$ Additionally, we use, 10,000 uniformly distributed points in $(x, t=0)$ as well as 5,000 points on each of $(x=0, t)$ and $(x=L_t+L_c, t)$ to enforce the initial and the two boundary conditions. Any graph-based framework such as Theano \cite{bergstra2010theano}, Tensorflow \cite{abadi2016tensorflow}, or PyTorch \cite{paszke2019pytorch}, or high-level Application Programming Interfaces (API) of these platforms, such as Keras \cite{chollet2015keras}, used in this work, can be employed to efficiently construct the network and perform the optimization. There are also APIs, such as DeepXDE \cite{lu2019deepxde} or SciANN \cite{HAGHIGHAT2021113552}, that are specifically designed for setting up and training PINN models. We design a fully connected neural network to have 7 hidden layers and 30 nodes per layer (5701 parameters in total) for $\mathcal{N}_\alpha$, shown in \cref{fig3}. For the temperature networks $\mathcal{N}_T^-$ and $\mathcal{N}_T^+$, \cref{eq22} and \cref{fig4}, again we pick 7 layers with 20 nodes per layer for each networks (5202 parameters in total). A mini-batch optimization with a batch size of 512 is used employing the Adam optimizer \cite{kingma2014adam}, which is based on the stochastic gradient descent. A learning rate scheduler is also used for the optimization starting from $10^{-3}$ and reducing by a factor of $0.5$ in case there are no improvement in the total loss values after 20 and 10 epochs for temperature and degree of cure, respectively. The analysis is performed for 10 iterations with 30 epochs for each of $\alpha$ and $T$ in each iteration, i.e. 300 total epochs of training for each of $\mathcal{N}_{\alpha}$ and $\mathcal{N}_{T}$. The adaptive loss weight algorithm is also applied to both $\mathcal{N}_{\alpha}$ and $\mathcal{N}_{T}$ where loss weights are updated at the beginning of each iteration of training of each network, rather than each each training epoch, for reducing the computational cost of training. The transfer learning is used for two case studies to demonstrate computational efficiency obtained in the training of the network. Also, a PINN surrogate model is presented by extending the input features of the neural networks to include the tool material heat transfer coefficient, $h_t$. All the cases are modeled on an Intel(R) Core(TM) i7-9700 CPU PC with 32 GB of RAM. \par

\begin{table}[t!]
\centering
    \begin{tabular}{ |l|c|c|c| } 
        \hline
            \multirow{2}{3cm}{\textbf{Case study}} & \textbf{Boundary condition}  & \textbf{Thickness} & \multirow{2}{3.5cm}{\textbf{Section number in the paper}}\\
            &  ($\text{K}$ or $\text{W}/\text{m}^2 \text{K}$)&  ($\text{mm}$)  &\\ 
            \hline
            \multirow{2}{3cm}{Case 1} & Prescribed temperature  & \multirow{4}{2cm}{$L_t=20$\\$L_c=30$} &  \multirow{4}{3cm}{Section \ref{sec:CaseStudies}} \\ 
            & $T_{bc_1} = T_a(t)$ and $T_{bc_2} = T_a(t)$  & &\\
            \cline{1-2}
            \multirow{2}{3cm}{Case 2} & Convective & & \\ 
            & $h_t = 70$ and $h_c = 120$  &   &\\
            \hline
            \multirow{2}{3cm}{Case 3} & Prescribed temperature  & \multirow{4}{2cm}{$L_t=200$\\$L_c=300$} &  \multirow{4}{3cm}{\ref{sec:appendix}} \\
            & $T_{bc_1} = T_a(t)$ and $T_{bc_2} = T_a(t)$    & &\\
            \cline{1-2} 
            \multirow{2}{3cm}{Case 4} & Convective & & \\ 
            & $h_t = 70$ and $h_c = 120$  & &\\
        \hline
    \end{tabular}
    \caption{Thickness and boundary conditions for different case studies.}
    \label{table4}
\end{table}

\subsection{Predictions of temperature and degree of cure}\label{predtempalpha}
\begin{figure}[t!]
    \centering
    \includegraphics[width=\textwidth,height=\textheight,keepaspectratio]{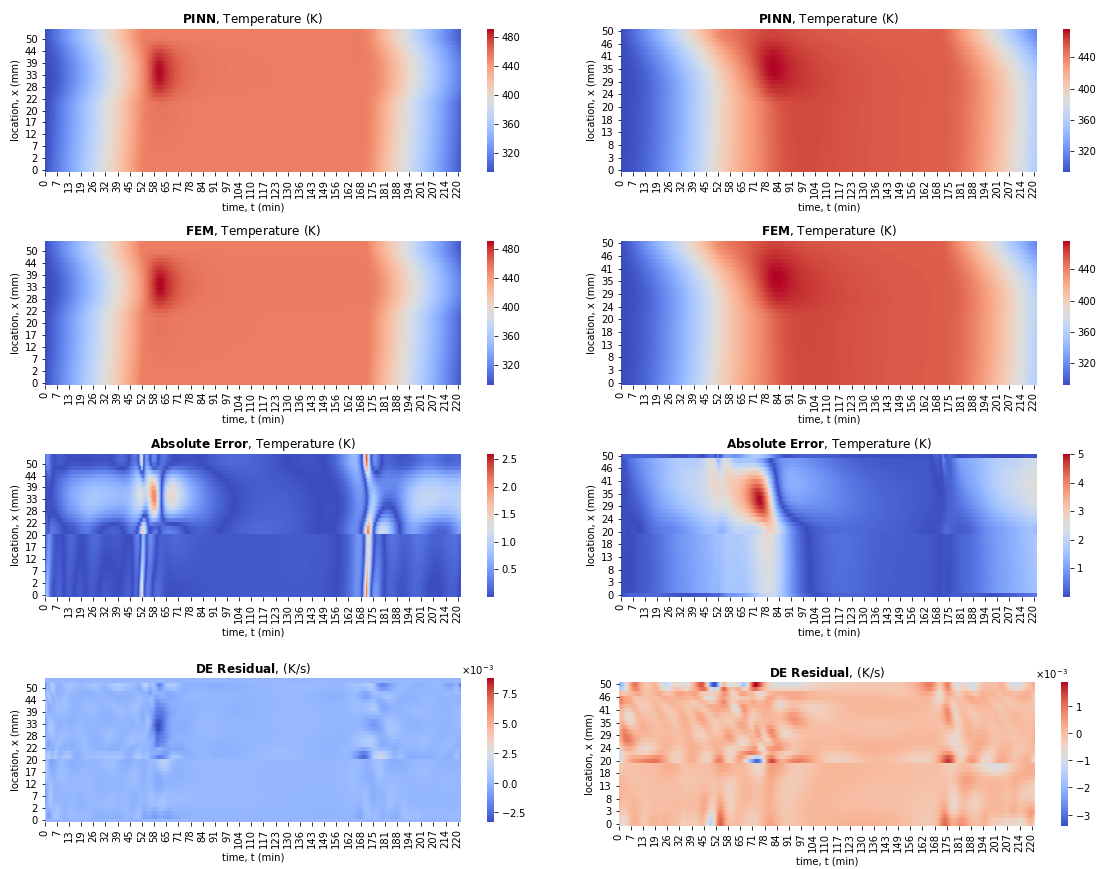}
    \caption{PINN and FEM temperature predictions for (Left) case study 1 and (Right) case study 2. (Top row) predicted temperature using PINN (Second top row) predicted temperature using FEM (Third top row) absolute error of the predicted temperature between PINN and FEM. (Bottom row) residual of temperature differential equation (\cref{eq3})} 
    \label{fig5}
\end{figure}

\begin{figure}[t!]
    \centering
    \includegraphics[width=\textwidth,height=\textheight,keepaspectratio]{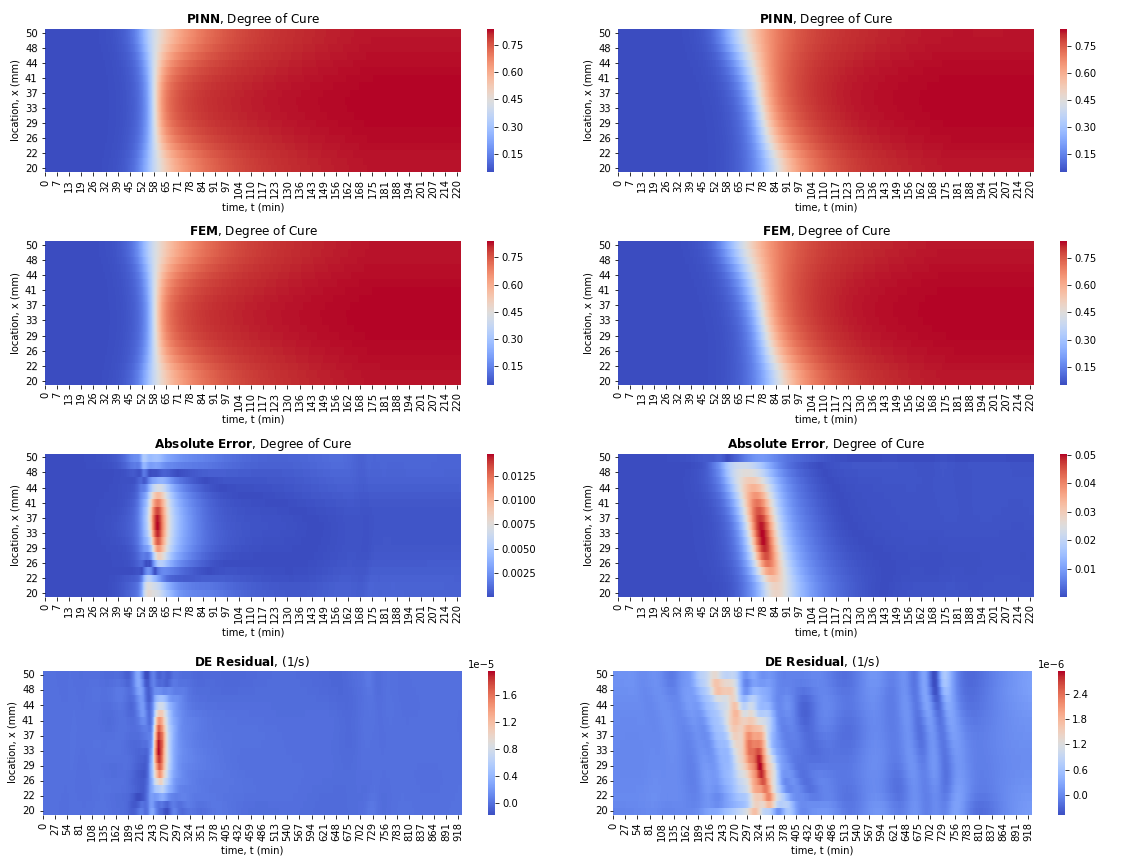}
    \caption{PINN and FEM degree of cure predictions for (Left) case study 1 and (Right) case study 2. (Top row) predicted degree of cure using PINN (Second top row) predicted degree of cure using FEM (Third top row) absolute error of the predicted degree of cure between PINN and FEM. (Bottom row) residual of degree of cure differential equation (\cref{eq4})}
    \label{fig6}
\end{figure}

\begin{figure}[t!]
    \centering
    \includegraphics[width=\textwidth,height=\textheight,keepaspectratio]{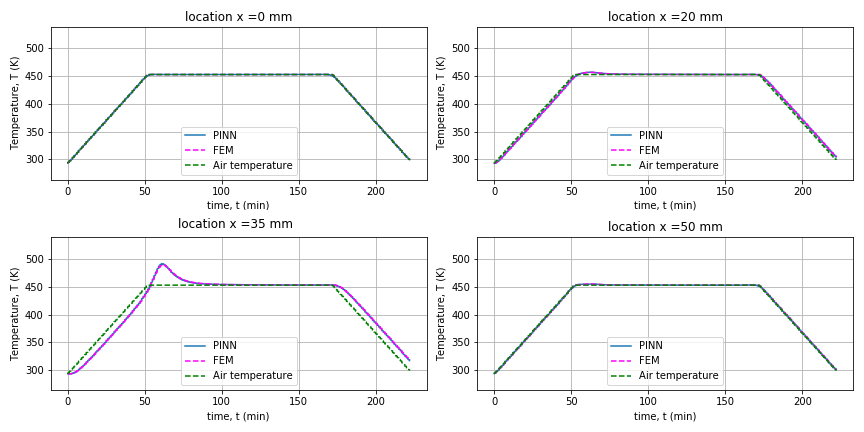}
    \caption{Temperature history for case study 1 (prescribed temperature at boundaries) at tool bottom boundary ($x=\SI{0}{\mm}$), tool-composite intersection ($x=\SI{20}{\mm}$), middle of composite ($x=\SI{35}{\mm}$), and composite boundary top ($x=\SI{50}{\mm}$).}
    \label{fig7}
\end{figure}

\begin{figure}[t!]
    \centering
    \includegraphics[width=\textwidth,height=\textheight,keepaspectratio]{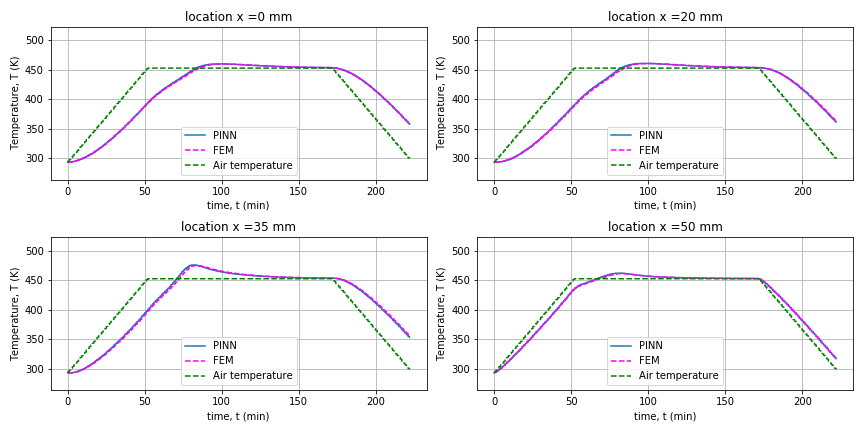}
    \caption{Temperature history for case study 2 (convective temperature at boundaries) at tool bottom boundary ($x=\SI{0}{\mm}$), tool-composite intersection ($x=\SI{20}{\mm}$), middle of composite ($x=\SI{35}{\mm}$), and composite boundary top ($x=\SI{50}{\mm}$).}
    \label{fig8}
\end{figure}

To check the accuracy of the PINN method in prediction of $T$ and $\alpha$ field variables, results are compared against FE predictions from RAVEN 3.13.1 \cite{raven3131} software. The FEM discretization includes 35 elements for case studies 1 and 2 and 260 elements for case studies 3 and 4, with 926 time-steps for all cases. The temperature and degree of cure predictions for case studies 1 and 2 are shown in \cref{fig5,,fig6}. The relative percentage error of the predicted temperature between PINN and FEM is less than 1\%. Also, the differential equations residuals errors (\cref{eq3,,eq4}) are small everywhere on the domain. For degree of cure, the location of maximum residual errors almost matches the location of maximum absolute error between PINN and FEM. In \cref{fig7,,fig8}, the temperature evolution at the bottom of the tool, composite-tool intersection, middle of the composite material, and top of the composite are shown for the full cure cycle. PINN accurately captures the maximum part temperature, i.e. \emph{exotherm}, that occurs at the middle of the composite part due to internal heat generated from resin reaction.  For case study 1, equality of the tool and composite temperature at the boundary of the autoclave air temperature is captured accurately. For the case study 2, the temperature at the boundaries lags behind the air temperature due to convective heat transfer, which the maximum difference between air and tool/part temperature known as \emph{thermal lag}.  Due to lower convective heat transfer coefficient, $h$, for the tool material, the thermal lag is more pronounced compared to the composite. Accurate predictions of both the exotherm and thermal lag are of great importance in processing of composite materials inside autoclaves. To further check the performance of the model, an extreme and unrealistic case involving $L_c=  \SI{300}{\mm}$ and $ L_t = \SI{200}{\mm}$ is also performed the results of which are presented in \ref{sec:appendix}. 

\FloatBarrier

\subsection{Effect of adaptive loss weights technique}\label{effectadaptiveloss}
\begin{figure}[t!]
    \centering
    \includegraphics[width=\textwidth,height=\textheight,keepaspectratio]{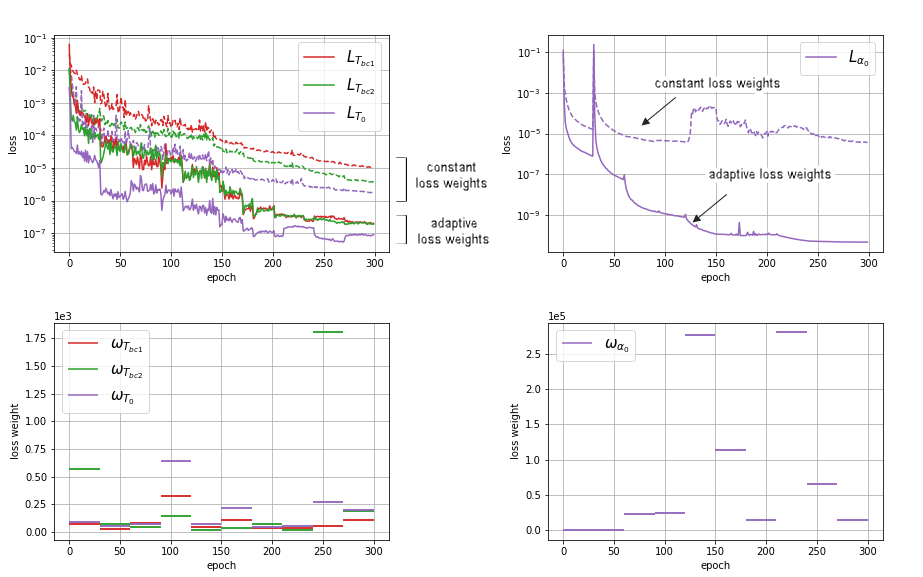}
    \caption{Evolution of initial and boundary conditions (Top) loss terms and (Bottom) loss weights for case study 1 for optimization with constant and adaptive loss weights (Left) temperature (Right) degree of cure. For the training with the adaptive loss weights, loss weights are updated at the beginning of each iteration. All the loss weights are one for the case of constant loss weights.}
    \label{fig9}
\end{figure}

\begin{figure}[t!]
    \centering
    \includegraphics[width=\textwidth,height=\textheight,keepaspectratio]{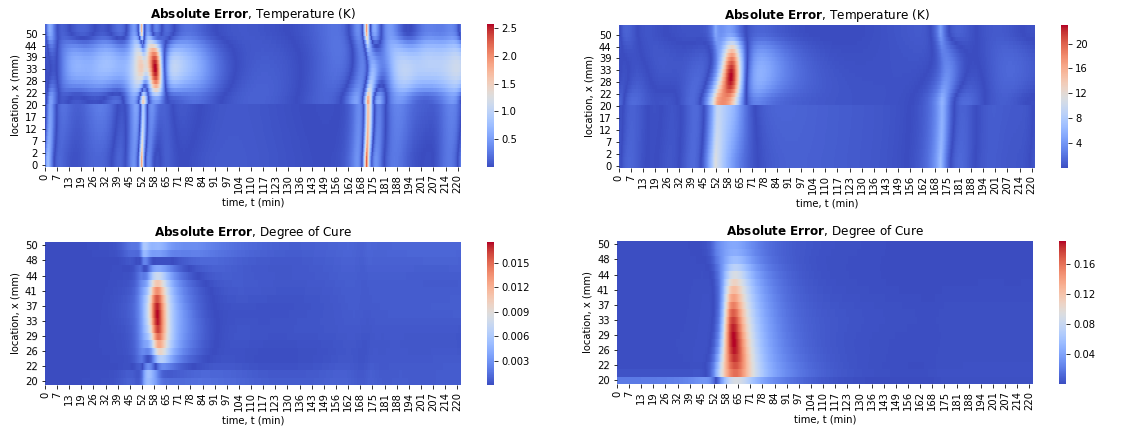}
    \caption{Effect of adaptive loss weight algorithm on error between PINN and FEM predictions for case study 1 (Left) training with adaptive loss technique (Right) training with constant loss weights.}
    \label{fig10}
\end{figure}

To highlight the importance of employing the adaptive loss weights presented in \cref{adaptiveloss}, the performance of the proposed PINN without such dynamic update of loss weights is presented here. For the case study 1, the evolution of losses and loss weights associated with initial and boundary conditions for both temperature and degree of cure are shown in \cref{fig9} where significantly better convergence is observed when the adaptive loss weight method is used. The error in temperature and degree of cure predictions are also shown in \cref{fig10} where significantly higher accuracy in the predictions is observed for the training with adaptive loss weight technique as a consequence of network balanced training. The error of prediction is about 8 and 10 times higher for temperature and degree of cure, respectively, if constant loss weights are used.

\FloatBarrier

\subsection{Effect of transfer learning on training time of PINN}\label{sec:trasnferlearningexample}
To demonstrate how transfer learning could improve computational efficiency in training of the PINN, an example is presented in this section where the training of a model with different material properties, i.e. conductivity of the tool and composite, is performed using transfer learning. Two new composite problems, similar to case studies 1 and 3, with a different tool and composite conductivity coefficients ${k_t}$ and  ${k_c}$, respectively, are considered as detailed in \cref{table5}. For transfer learning, the trained weight matrices and bias vectors of the neural networks for case studies 1 and 3 are used as initial values of weights and biases of problems with similar geometry and boundary conditions, i.e. cases 1a and 3a respectively.  

\begin{figure}[ht!]
    \centering
    \includegraphics[width=\textwidth,height=\textheight,keepaspectratio]{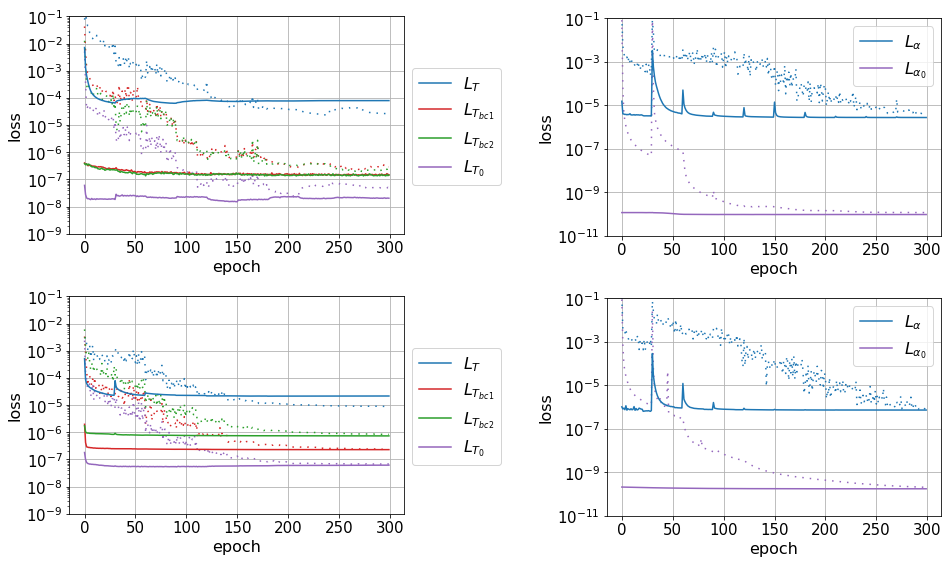}
    \caption{Evolution of loss terms for (Left) temperature and (Right) degree of cure for (Top) case studies 1 and 1a and (Bottom) case studies 3 and 3a. Dashed lines show the losses for the original training (case studies 1 and 3) and solid lines are for case studies with transfer learning (1a and 3a).}
    \label{fig14}
\end{figure}

The evolutions of different loss terms for all cases are shown in \cref{fig14} in comparison to loss terms during the training of their original problem. It is clear that the pre-trained PINN using transfer learning converges much faster than the original case. This is an immediate consequence of using transfer learning. The computation times for training of all cases are reported in \cref{table5}. 

\begin{table}[t!]
\centering
    \begin{tabular}{ |l|c|c|c|c|c| }
        \hline
            \textbf{Case study} & \textbf{Thickness (mm)} & \multicolumn{2}{c|}{\textbf{Conductivity ($\text{W}/\text{m K}$)}} & \textbf{Training time (s)} \\ 
            \textbf{} &\textbf{} & tool (${k_t}$) & composite (${k_c}$) & \\
        \hline
            Case 1 & 50 & 13.0 & 0.639 & 2370 (original training) \\
            Case 1a & 50 & 11.70 & 0.702 & 45 (transfer learning) \\ 
            \hline 
            Case 3 & 500 & 13.0 & 0.639 & 2390 (original training)\\
            Case 3a & 500 & 11.70 & 0.702 & 43 (transfer learning) \\
         \hline
    \end{tabular} 
    \caption{Simulation time for original training and training using transfer learning for different case studies. The trained network parameters of case studies 1 and 3 are used as initial values for the training of case studies 1a and 3a, respectively, with an early stopping during the training.}
    \label{table5}
\end{table}

\subsection{Prediction of temperature and degree of cure using PINN surrogate model}\label{sec:surrogatemodelingexample}
In this section we showcase the potential application of our developed PINN architecture as a physics-informed surrogate model of the composite-tool system discussed above. Let us assume that we would like to predict the response variables $T$ and $\alpha$ as a function of tool heat transfer coefficient variable, $h_t$, in addition to the space and time variables, i.e. $\boldsymbol{x}:=(x,t, h_t)$; the $h_t$ variable is added as an input to the network architectures, i.e. $\zeta=h_t$ in \cref{fig15}. The training of the surrogate model is conducted on the same $x$ and $t$ points inside and on the boundaries of the domain as detailed in \cref{sec:CaseStudies} for case study 3. While we can add random collocation points for $h_t$ and train this model without any labeled data, to accelerate the training, in addition to the physical constraints introduced earlier, we also add three data sets associated with $h_t={50, 60, 70}$. For simplicity, all the network and optimization hyper-parameters are the same as the original PINN model used for previous case studies.

After training of PINN surrogate model, we can evaluate $T$ and $\alpha$
for any given values of $h_t$ within the considered range. This takes about
$1.5 \times 10^{-5}$ seconds per evaluation; and for a grid including 34 points for
location and 926 points for time, this amounts to a total of about 0.5 seconds.
For comparison, solving the same problem using FEM takes about 5 seconds.
Therefore if one is interested in the solution only at a particular
spatio-temporal point, this represents a 5-order-of-magnitude reduction in
computation time; and even if one is interested in the solution on the $34
\times 926$ grid, this still provides an order-of-magnitude reduction. 

\begin{figure}[ht!]
    \centering
    \includegraphics[width=\textwidth,height=\textheight,keepaspectratio]{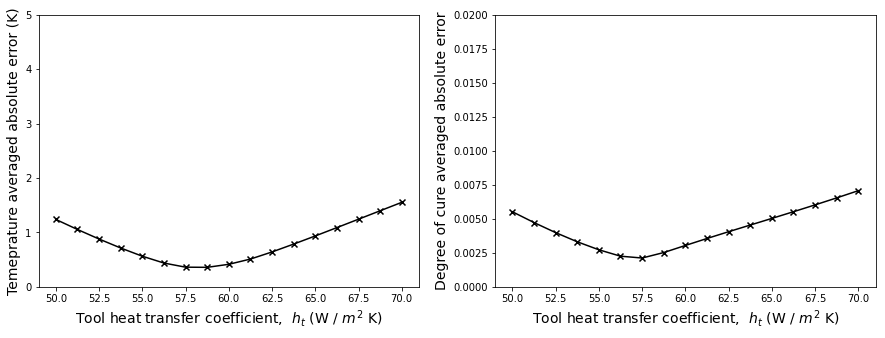}
    \caption{Averaged absolute error between prediction from PINN surrogate model and FEM for different values of tool heat transfer coefficient $h_t$ (Left) temperature (Right) degree of cure.}
    \label{fig16}
\end{figure}

\begin{figure}[ht!]
    \centering
    \includegraphics[width=\textwidth,height=\textheight,keepaspectratio]{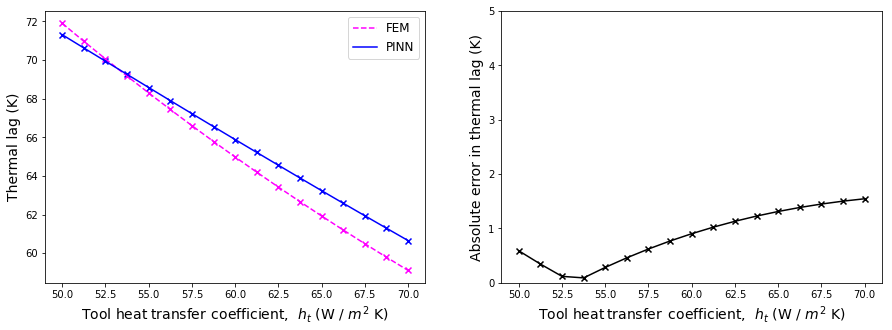}
    \caption{Prediction of thermal lag from PINN surrogate model and FEM (Left) thermal lag value (Right) absolute error between thermal lag values predicted from two models.}
    \label{fig17}
\end{figure}

The averaged absolute error of the predicted
temperature and degree of cure from the PINN surrogate model and FEM are shown
in \cref{fig16}. We find that the averaged absolute error remains small
for all cases, with less than 1.6 K for the temperature and less than 0.007 for
the degree of cure. These error values are well within the $5$ K
acceptable/measurable range of error for temperature and $0.02$ for degree of
cure for processing of composite materials. In processing of composite
materials, the $h_t$ mostly affects the thermal lag (see \cref{predtempalpha}
for the definition). In \cref{fig17}, the predicted thermal lag from
both models are plotted where the absolute error in prediction of thermal lag
is less than $1.6$ K. The case study presented here is an indication of
potential of the PINN approach as a surrogate model in more complex cases;
although results are within an acceptable range for practical application,
there is certainly room for improvement and exploration that we leave for future research.

 \FloatBarrier

\section{Conclusions}\label{sec:Conclusion}

In this study, we present a PINN framework for modelling of exothermic heat transfer in a composite-tool system undergoing a full cure-cycle. The proposed PINN is a feed-forward neural network that tackles solution of the two coupled differential equations, namely, the exothermic heat transfer and resin reaction which govern the distribution and evolution of temperature and degree of cure in the composite and tool materials. \par

A disjointed network architecture is employed to utilize independent networks
parameters for the output field variables, i.e. temperature and degree of cure.
A sequential approach for training of the proposed disjointed PINN is presented
that overcomes instability in the training of the PINN. In the training
process, the network parameters are constrained through introduction of
physics-based loss functions. Additionally, the standard PINN is modified to
capture discontinuities of solution variables and/or their derivatives as a
consequence of material discontinuities in the solution domain. The continuity
of physical quantities, such as heat flux, are maintained using additional
physics-based loss terms. We also employed the adaptive loss weight algorithm
based on the scaling gradient of different terms for a robust training of the
neural network. We showed how transfer learning could be applied to improve training
of the PINN, and demonstrated its extension 
to the surrogate modeling setting by including the heat transfer coefficient
as an input parameter. Comparisons between
the numerical results from PINN and those obtained from classical numerical
methods demonstrate the accuracy of the proposed methodology,
as well as significant gains in computation time in the surrogate setting. \par

The proposed PINN does not require large amount of labeled data generated in
advance using high fidelity simulation tools. It also does not rely on the
domain discretization which is typically done in the classical numerical
methods. Such characteristics along with the accuracy in predictions of
solution variables offers the possibility of using the proposed PINN for
surrogate modeling of composite materials processing as well as probabilistic
modeling, optimization, uncertainty quantification, and real-time monitoring
during processing.

\section*{Acknowledgments}\label{sec:acknowledgments}

The authors would like to acknowledge the financial support for this research provided by the Natural Sciences and Engineering Research Council of Canada (NSERC) through Alliance grant (file number ALLRP 549167–19) with Convergent Manufacturing Technologies Inc. (CMT) as the industrial partner.

\appendix
\section{An unrealistically thick composite-tool example}\label{sec:appendix}

\begin{figure}[ht!]
    \centering
    \includegraphics[width=\textwidth,height=\textheight,keepaspectratio]{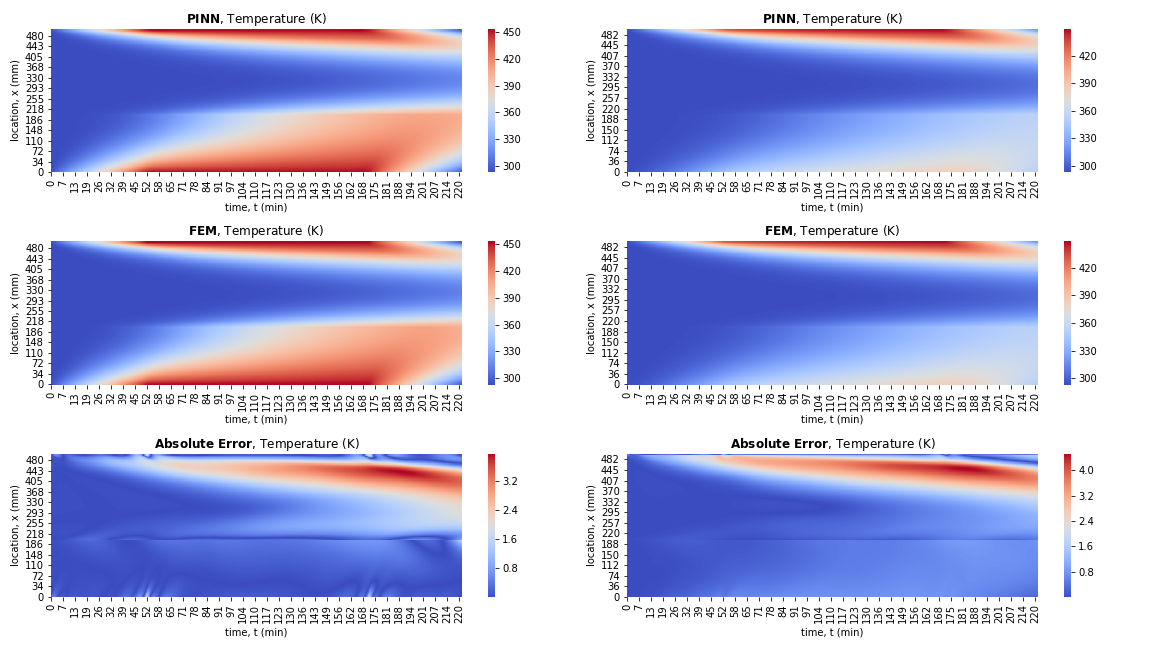}
    \caption{PINN and FEM degree of cure predictions for (Left) case study 3 and (Right) case study 4.}
    \label{fig11}
\end{figure}

\begin{figure}[ht!]
    \centering
    \includegraphics[width=\textwidth,height=\textheight,keepaspectratio]{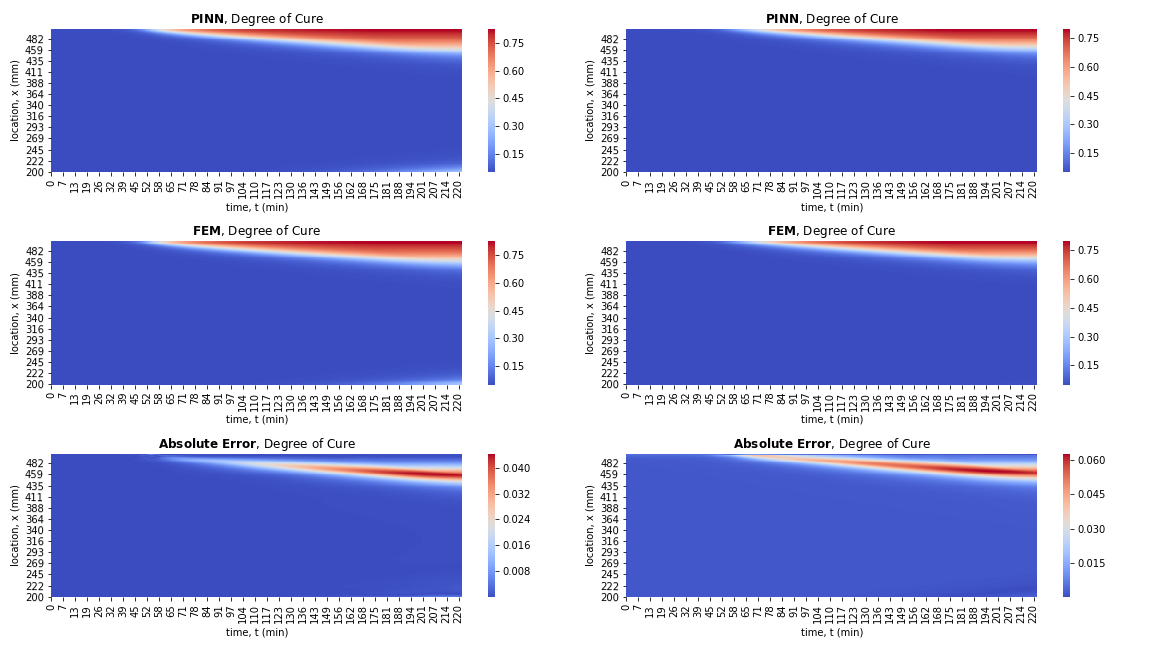}
    \caption{PINN and FEM degree of cure predictions for (Left) case study 3 and (Right) case study 4.}
    \label{fig12}
\end{figure}

\begin{figure}[ht!]
    \centering
    \includegraphics[width=\textwidth,height=\textheight,keepaspectratio]{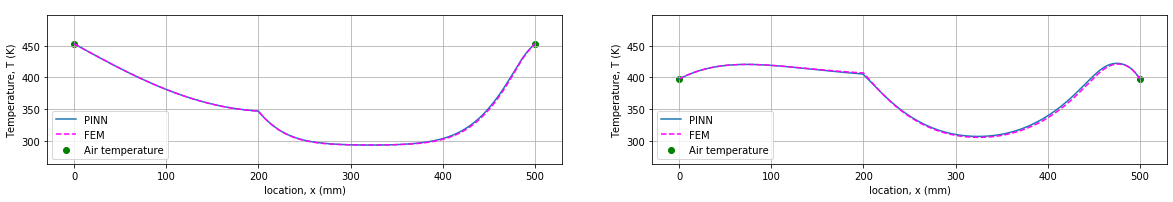}
    \caption{Temperature distribution for case study 3 (prescribed temperature at boundaries) (Left) at $t = \SI{90}{min}$ and (Right) at time $t = \SI{190}{min}$.}
    \label{fig13}
\end{figure}

The distribution of temperature $T$ and degree of cure $\alpha$ is affected significantly by the size of composite and tool. To demonstrate the accuracy of the proposed PINN for a wide range of material size, results of the exothermic heat transfer model an on an extreme and unrealistic composite-tool system with $L_c = \SI{300}{\mm}$ and $L_t = \SI{200}{\mm}$ corresponding to the case studies 3 and 4 in \cref{table4} are presented here. The temperature and degree of cure are shown in \cref{fig11,,fig12} where again a good match between PINN predictions and those of FEM from RAVEN software is observed. To highlight the accuracy of the PINN, the temperature through the thickness of the tool and composite is shown in \cref{fig13} where the extreme and high discontinuity of temperature spatial derivative is observed. Such performance is a direct result of neural network decomposition at the composite-tool intersection point as discussed in \cref{bimaterial} and shown in \cref{fig4}.

\FloatBarrier

\clearpage
\bibliographystyle{unsrtnat}
\biboptions{sort&compress}

\bibliography{sample}

\end{document}